%% file: main.tex
\documentclass{sigchi-ext}
\usepackage{xspace}
\usepackage[T1]{fontenc}
\usepackage{textcomp}
\usepackage[scaled=.92]{helvet} 
\usepackage{graphicx} 
\usepackage{balance}  
\usepackage{booktabs} 
\usepackage{ccicons}  
\usepackage{ragged2e} 
\usepackage{color,soul}
\usepackage{subcaption}
\usepackage{enumitem}

\setul{0.5ex}{0.3ex}

\input{includes.tex}
\usepackage{xcolor}
\definecolor{RoyalBlue}{rgb}{0.36, 0.54, 0.66}
\colorlet{lightmintbg}{RoyalBlue!40}
\colorlet{lightermintbg}{RoyalBlue!50}
\colorlet{lightestmintbg}{RoyalBlue!60}

\def\plaintitle{RECAST: Interactive Auditing of Automatic Toxicity Detection Models} 
\def\plainauthor{Austin P Wright, Omar Shaikh, Haekyu Park, Will Epperson, Muhammed Ahmed, Stephane Pinel, Diyi Yang, Duen Horng (Polo) Chau}

\def\plainkeywords{Machine Learning Fairness; Interactive Visualization; Algorithmic Bias; Natural Language Processing}

\title{RECAST: Interactive Auditing of Automatic Toxicity Detection Models}

\numberofauthors{8}
\author{%
  \alignauthor{%
    \textbf{Austin P Wright}\\
    \affaddr{Georgia Institute of Technology} \\
    \email{apwright@gatech.edu} }
  \alignauthor{%
    \textbf{Muhammed Ahmed}\\
    \affaddr{Mailchimp}\\
    \email{muhammed.ahmed@mailchimp.com} } \vfil 
  \alignauthor{%
    \textbf{Omar Shaikh}\\
    \affaddr{Georgia Institute of Technology}\\
    \email{oshaikh@gatech.edu} }
  \alignauthor{%
    \textbf{Stephane Pinel}\\
    \affaddr{Mailchimp}\\
    \email{stephane.pinel@mailchimp.com} } \vfil
  \alignauthor{%
    \textbf{Haekyu Park}\\    
    \affaddr{Georgia Institute of Technology}\\
    \email{haekyu@gatech.edu} }
  \alignauthor{%
    \textbf{Diyi Yang}\\
    \affaddr{Georgia Institute of Technology}\\
    \email{diyi.yang@cc.gatech.edu} } \vfil
  \alignauthor{%
    \textbf{Will Epperson}\\
    \affaddr{Georgia Institute of Technology}\\
    \email{willepp@gatech.edu} } 
  \alignauthor{%
    \textbf{Duen Horng Chau}\\
    \affaddr{Georgia Institute of Technology}\\
    \email{polo@gatech.edu} } 
}

\definecolor{linkColor}{RGB}{6,125,233}

\hypersetup{%
  pdftitle={\plaintitle},
  pdfauthor={\plainauthor},
  pdfkeywords={\plainkeywords},
  bookmarksnumbered,
  pdfstartview={FitH},
  colorlinks,
  citecolor=black,
  filecolor=black,
  linkcolor=black,
  urlcolor=linkColor,
  breaklinks=true,
}

\setcopyright{rightsretained} 
 
\begin{document}

\CopyrightYear{2020} 
\setcopyright{rightsretained} 
\conferenceinfo{Chinese CHI 2020}{April  26, 2020, Honolulu, HI, USA}
\doi{10.1145/3403676.3403691}
\isbn{978-1-4503-8815-3/20/04}
\copyrightinfo{\acmcopyright}

\maketitle

\RaggedRight{} 

\begin{abstract}
    As toxic language becomes nearly pervasive online, there has been increasing interest in leveraging the advancements in natural language processing (NLP) to automatically detect and remove toxic comments. Despite fairness concerns and limited interpretability, there is currently little work for auditing these systems in particular for end users. We present our ongoing work, \tool{}, an interactive tool for auditing toxicity detection models by visualizing explanations for predictions and providing alternative wordings for detected toxic speech. \tool{} displays the attention of toxicity detection models on user input, and provides an intuitive system for rewording impactful language within a comment with less toxic alternative words close in embedding space. Finally we propose a larger user study  of \tool{}, with promising preliminary results, to validate it's effectiveness and useability with end users.
\end{abstract}

\keywords{\plainkeywords}

\ccsdesc[500]{Human-centered computing~Visualization application domains}

\printccsdesc

\section{Introduction}
There is a growing desire to moderate and remove toxic language from social media and public forums, as a result of increasing online interactions \cite{fortuna2018survey}. For example in a 2015 user survey of the online social network platform reddit, 50\% of people who wouldn't’t recommend reddit cited hateful or offensive content and community as the reason why, however on the other hand 35\% of complaints from extremely dissatisfied users were about heavy handed moderation and censorship\cite{buni_chemaly_2016}. This balance of issues is further complicated by the high volume of interaction on these platforms, making manual moderation often not tractable and leading to the development of automatic toxicity detection models such as the Google Perspective API \cite{perspectiveAPI}. 
 
However, 
the existing body of research highlights potential flaws in the underlying language models for toxicity detection systems.
For example, several word embedding models and their training language datasets exhibit biases towards certain subgroups such as gender and race~\cite{bolukbasi2016man, sap-etal-2019-risk}.
Some widely deployed NLP models, specifically BERT, also tend to overlook simple linguistic structures like negation, reducing their effectiveness~\cite{ettinger2019bert}.
A serious problem is that it is difficult to fix the potential erroneous model outputs 
due to a lack of the interpretability of NLP models.

\begin{marginfigure}
  \begin{minipage}{\marginparwidth}
    \centering
    \includegraphics[width=1\marginparwidth]{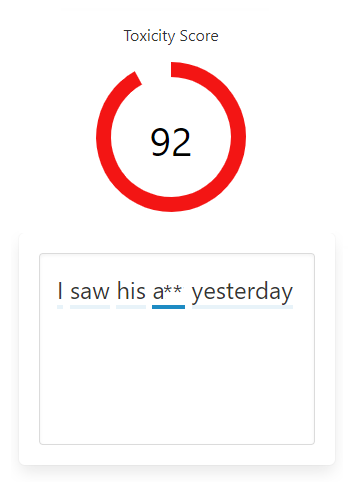}
    \caption{A case where a user might notice and flag biases in their model with respect to dialects. The language shown above is non-toxic in the African American English (AAE) dialect, yet \tool shows a fairly high toxicity score.}
    \label{aae-toxic}
  \end{minipage}
\end{marginfigure}

These flaws in language models persist in toxic statement detection systems, especially for end users. 
Therefore users of online forums that use toxicity detection systems based on black-box NLP models might question how their language is being examined. 
Nonnative speakers may inadvertently write language that could confuse the system and be marked as toxic. 
Without tools designed for actual end users to audit what is being detected and make actionable changes to their language, people are disempowered to participate in discourse online. 
Furthermore, without an ability to detect when a model is falsely flagging language due to either linguistic limitations or social biases, the work of finding and correcting bias are left entirely to the unrepresentative population of machine learning researchers and software engineers. 
Therefore \textbf{the ability to audit these models must be provided to end-users affected by them.}
\section{\tool}

We address these challenges by developing an interactive tool called \textbf{\tool{}}, which allows for the interrogation of toxicity detection models through counterfactual alternative wording and attention visualization. This design does not require any expertise in machine learning, but enables users to visualize their sentence through the eyes of the algorithm.  \tool{} currently supports analysis of a fine-tuned BERT model on the Jigsaw Toxicity dataset \cite{jigsaw}. Our ongoing work presents the following contributions and vision:

\begin{figure}
  \centering
  

  \hspace*{-0.6\columnwidth}
  \parbox{1.45\columnwidth}{
    
    \includegraphics[width=1.35\columnwidth]{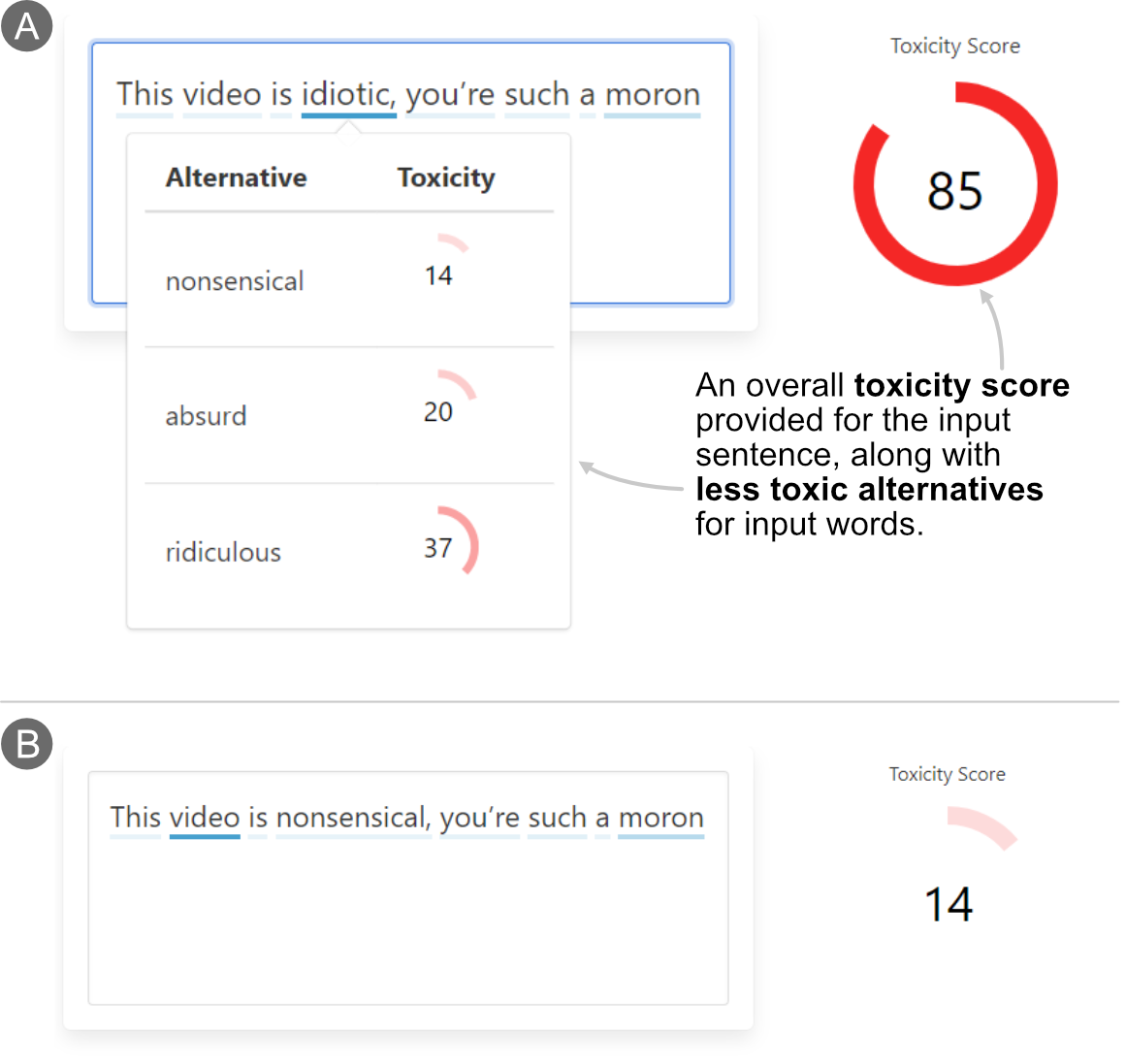}
    
      \caption{\textbf{A}: \tool{} consists of a textbox and a radial progress bar. A color change on the radial progress, along with a score, indicate the toxicity of a sentence. Toxicity ranges from white (non-toxic) to red (very toxic). Users can hover over options to preview toxicity scores for replacing the selected word in the sentence. 
      \textbf{B}: upon replacing the word (in the case of this figure, replacing ``idiotic'' with ``nonsensical''), the main radial progress bar reflects the reduced toxicity score. However, the small attention on the other pejorative word "moron," compared to "video" in the alternative version, shows the idiosyncrasies of the model and underlying dataset.}
      \label{overview-fig}

    }

\end{figure}

\tool is an interactive system (Figure \ref{overview-fig}) that allows users to input text and view the toxicity of the overall sentence, along with which words most contribute to output score. 
\tool displays a score between 0 and 100, which represents the probability that the model will classify the input as toxic. Users can view highly attended words, edit the text, select suggested alternate wordings, and watch the toxicity score dynamically update. Its design is purposefully simple, mirroring the text interaction techniques of underlining to note where editing is required and selecting listed alternatives which end users are familiar with from common software like Microsoft Word or Grammarly. This accessibility allows RECAST to effectively communicate the complexities of toxicity detection models using a visual language users are already fluent in.

\subsection{Attention Visualization}
We use attention to explain which words affect our model's choices. 
Various visualisation concepts can also be used to show the relative importance of words, such as highlighting and text opacity \cite{attn_vis_notebook}. However, we utilized an underline on every word, where the opacity of each underline would be controlled by attention placed on each word. We found that using an \setulcolor{RoyalBlue}\ul{underline} instead of adjusting the \textcolor{lightgray}{opacity} or \hl{highlighting} the word helped with legibility of the text, which is vital for users understanding differences in textual classifications.

\subsection{Alternative Wording}
Alternative wording provides users with options to swap or delete words in a sentence that are responsible for high toxicity scores. Figure \ref{overview-fig} highlights such a use of \tool. The underline visualization draws the user's attention to the most impactful words. When the user hovers over these words, suggested substitutions are shown and ranked by using the k-nearest words from Word2Vec embeddings \cite{10.5555/2999792.2999959}. Selecting one of these alternatives replaces the word and the new toxicity score is displayed to the right. This mode of interaction is easy and intuitive for users due to its similarity to familiar spellcheck or thesaurus tools and requires little retyping of edits. Furthermore it displays a range of options, which allows the end user agency in maintaining the original meaning as closely as possible. Finally, beyond the act of making the sentence less toxic, the technique allows users to learn which words tend to be highlighted, and what common synonyms the algorithm tends to suggest. This allows people to learn about the model and use this knowledge while writing future comments.

\section{Vision for Future Work}
Since the purpose of \tool{} is to provide power to end-users, an important feature to include is an ability to flag when the model gets it wrong. These examples can be used to provide researchers with data for retraining their models, and provide an avenue of recourse for people adversely affected by the errors in the model. Therefore the statements about the accessibility and usability of \tool{} must be validated empirically by a user study. To this end, we plan on evaluating end-users' capacity to reduce toxicity in a sample text given \tool{}. In preparation for a full study we have run a small pilot on 18 participants through Amazon Mechanical Turk with approval from the Institutional Review Board, where we found that users given \tool{} rated the usefulness of the tool in reducing toxicity on a 5 point scale (higher being more useful) an average of 4.4, compared to 3.6 for a control group. While these results are not highly statistically significant, they help justify further exploration into this tool.

\section{Conclusion}
\tool takes steps towards increased transparency for black-box NLP models that are responsible for moderating large swaths of the internet. By enabling users to interact with text input, view alternative wordings for toxic sentences, and identify potential biases, \tool provides insights about models to those people actually affected by them, and allows everyone to participate online in both a \textbf{less toxic}, and \textbf{more fair}, environment.

\section{Acknowledgements}
This work was supported in part by NSF grants IIS-1563816, CNS-1704701, NASA NSTRF, gifts from Intel (ISTC-ARSA), NVIDIA, Google, Symantec, Yahoo! Labs, eBay, Amazon.


\bibliographystyle{SIGCHI-Reference-Format}
\bibliography{main}

\end{document}

%% file: includes.tex
\newcommand{\tool}[0]{\textsc{Recast}\xspace}

\definecolor{orange}{RGB}{215,87,0}
\definecolor{red}{RGB}{220,0,0}
\definecolor{agreen}{RGB}{74, 198, 148}
\definecolor{purple}{RGB}{173, 141,174}
\definecolor{aqua}{RGB}{87, 180, 181}

\newcommand{\todo}[1]{\textcolor{red}{[#1]}}
\newcommand{\polo}[1]{\textcolor{red}{[#1 -Polo]}}
\newcommand{\diyi}[1]{\textcolor{blue}{[#1 -Diyi]}}
\newcommand{\austin}[1]{\textcolor{agreen}{[#1 -Austin]}}
\newcommand{\omar}[1]{\textcolor{purple}{[#1 -Omar]}}
\newcommand{\haekyu}[1]{\textcolor{orange}{[#1 -Haekyu]}}
\newcommand{\will}[1]{\textcolor{orange}{[#1 -Will]}}
\newcommand{\muhammed}[1]{\textcolor{agreen}{[#1 -Fred]}}
\newcommand{\stephane}[1]{\textcolor{aqua}{[#1 -Robert]}}

\newcommand{\mkclean}{
   \renewcommand{\todo}[1]{}
   \renewcommand{\austin}[1]{}
   \renewcommand{\haekyu}[1]{}
   \renewcommand{\diyi}[1]{}
   \renewcommand{\polo}[1]{}
   \renewcommand{\omar}[1]{}
   \renewcommand{\will}[1]{}
   \renewcommand{\muhammed}[1]{}
   \renewcommand{\stephane}[1]{}

}
\mkclean